\documentclass[10pt,twocolumn,letterpaper]{article}

\usepackage{iccv}
\usepackage{times}
\usepackage{epsfig}
\usepackage{graphicx}
\usepackage{amsmath}
\usepackage{amssymb}
\usepackage{algorithm}
\usepackage{algorithmicx}
\usepackage{algpseudocode}
\usepackage{appendix}
\usepackage{caption}
\usepackage{subcaption}
\usepackage{multirow}
\usepackage{booktabs}
\usepackage{color}

\usepackage[pagebackref=true,breaklinks=true,letterpaper=true,colorlinks,bookmarks=false]{hyperref}

\iccvfinalcopy 

\def\iccvPaperID{2144} 

\newcommand{\sethree}{$\mathbb{SE}\left( 3 \right)$}
\newcommand{\sethreeAlg}{$\mathfrak{se}\left( 3 \right)$}
\newcommand{\capOper}{\Psi }

\ificcvfinal\fi
\begin{document}

\title{Joint Estimation of Camera Pose, Depth, Deblurring, and Super-Resolution from a Blurred Image Sequence}

\author{Haesol Park \qquad  Kyoung Mu Lee\\
Department of ECE, ASRI, Seoul National University, 151-742, Seoul, Korea\\
{\tt\small haseol.park@gmail.com,  kyoungmu@snu.ac.kr}
}

\maketitle

\begin{abstract}
The conventional methods for estimating camera poses and scene structures from severely blurry or low resolution images often result in failure. The off-the-shelf deblurring or super-resolution methods may show visually pleasing results. However, applying each technique independently before matching is generally unprofitable because this na\"{\i}ve series of procedures ignores the consistency between images. In this paper, we propose a pioneering unified framework that solves four problems simultaneously, namely, dense depth reconstruction, camera pose estimation, super-resolution, and deblurring. By reflecting a physical imaging process, we formulate a cost minimization problem and solve it using an alternating optimization technique. The experimental results on both synthetic and real videos show high-quality depth maps derived from severely degraded images that contrast the failures of na\"{\i}ve multi-view stereo methods. Our proposed method also produces outstanding deblurred and super-resolved images unlike the independent application or combination of conventional video deblurring, super-resolution methods.


\end{abstract}

\section{Introduction}
\label{sec:introduction}
Structure from motion or multi-view stereo (MVS) is a very interesting problem in computer vision that aims to determine the underlying 3D scene structure and camera configuration from multiple images.
Despite the inherent difficulty of this inverse problem, contemporary algorithms show a satisfactory performance when applied on public datasets~\cite{middlebury2005, middlebury2003}. 

\begin{figure}[t]
\centering
	\includegraphics[width=0.475\textwidth]{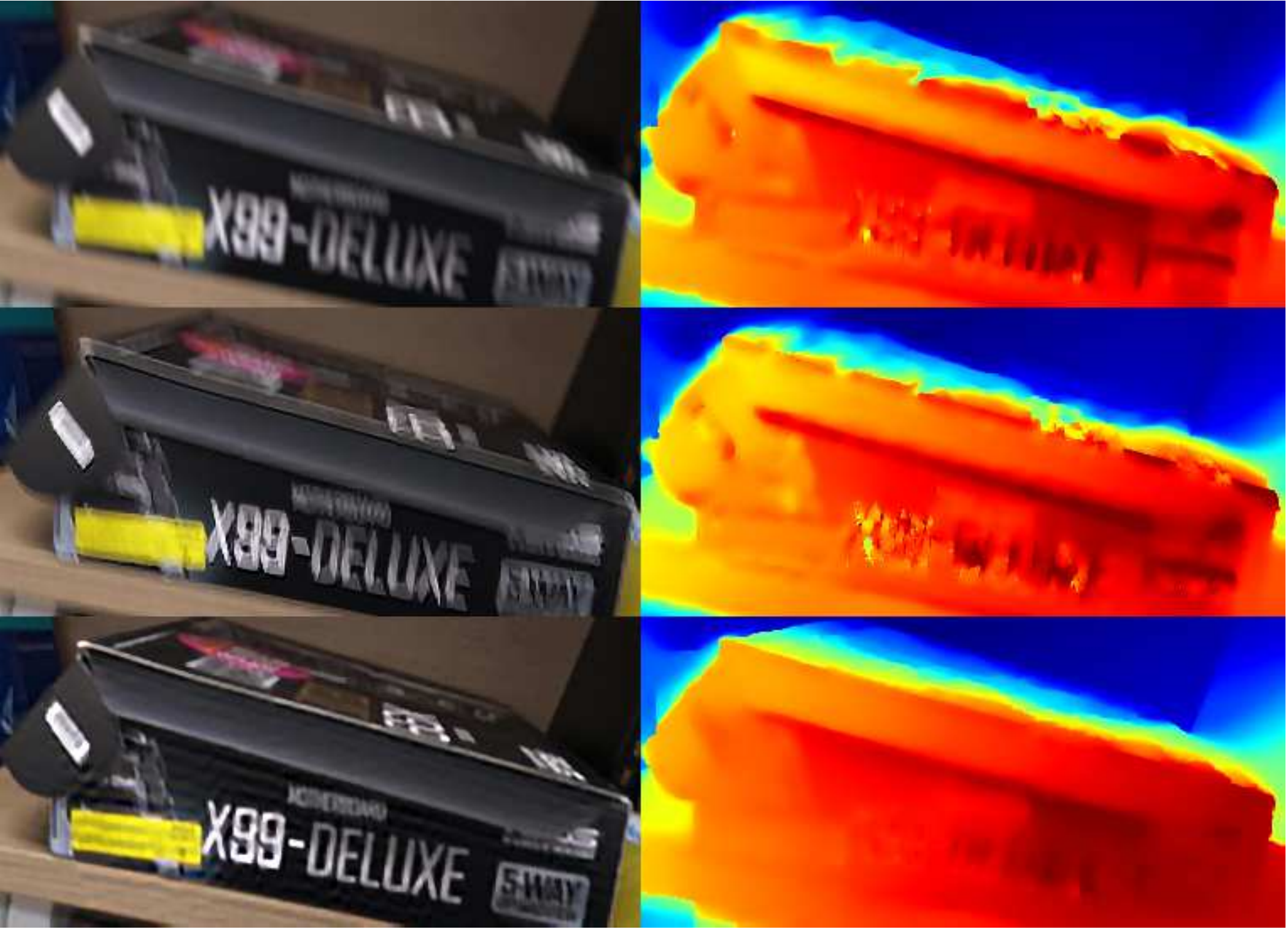}
	\caption{Comparison of depth estimation and image restoration results for blurry, low-resolution images. The left column shows the estimated latent images, while the right column shows their corresponding depth maps. The images from top to bottom are obtained via (a) a simple bicubic interpolation, (b) the independent use of deblurring~\cite{Jia2013} after applying the super-resolution algorithm~\cite{wang2015deep}, and (c) the proposed method, respectively. The depth maps for the first two rows are estimated via baseline variational depth estimation.}
\label{fig:indepVSoriVSours}
\end{figure}

\begin{figure*}[t]
\begin{center}	
   \includegraphics[width=0.98\linewidth]{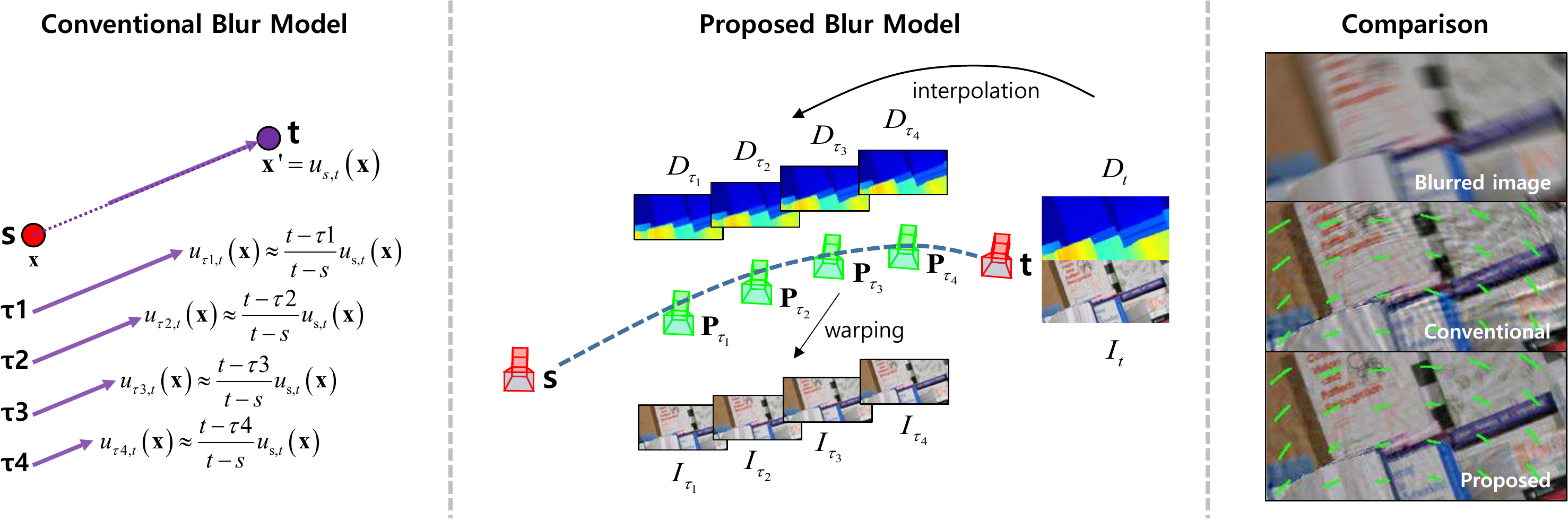}
\end{center}
   \caption{Comparison of the proposed blur model and the conventional blur model used in~\cite{Lee2013, Portz2012}. Both models illustrate the blur procedures for the frame at time $t$, where $s$ represents the time of the previous frame. The proposed model approximates the intermediate images $I_{\tau}$\rq{}s during the shutter time using the interpolated camera poses ${{\bf{P}}_{{\tau}}}$\rq{}s and depth maps ${D_{{\tau}}}$\rq{}s, while the conventional model depends on a single optical flow map from $s$ to $t$, $u_{s,t}$ (e.g. ${u_{{\tau _1},t}}$ is used to approximate ${I_{{\tau _1}}}$). The deblurred images with the overlaid blur kernels of each model are also presented for comparison. Although both images are obtained using the ground-truth depth map and camera poses, the image obtained using the conventional blur model exhibits more artifacts because of inaccurate blur kernels.}
\label{fig:blurModels}
\end{figure*}

Despite their encouraging achievements, some limitations prevent the aforementioned methods from being applied in highly realistic scenarios. 
Among these challenges are the blurs resulting from camera motion~\cite{Lee2013,Portz2012}, which becomes serious when using handheld cameras or cameras attached to moving vehicles.
Blur operation acts differently on each pixel according to the scene depth and camera motion, and it breaks brightness constancy assumption among consecutive frames.

Low-resolution (LR) input images also often affect stereo matching accuracy~\cite{Bhavsar2010,Lee2013_2}, because low-quality cameras are frequently used considering the limitations in cost or space for some applications.
However, even in a high-resolution (HR) image, the actual scene resolution is spatially uneven and dependent on depth because of the perspective projection. 

The aforementioned problem becomes especially challenging when the image frames are simultaneously corrupted by blur and low resolution.
This problem can be directly addressed by applying the super-resolution~\cite{timofte2014} or deblurring method~\cite{Jia2013} before matching, which may produce visually pleasing images. 
However, the results obtained using this approach are worse than those obtained using original images in terms of matching as shown in Figures~\ref{fig:indepVSoriVSours} (a) and (b), because single-image super-resolution or deblurring algorithms ignore and break the brightness constancy among neighboring frames.

In this study, we consider the four inter-related problems of camera pose estimation, dense depth reconstruction, deblurring, and super-resolution as a whole by formulating them as a unified energy function. 
To the best of our knowledge, this study is the first to solve the four aforementioned problems jointly in a single framework. Our proposed method clearly outperforms the independent use of existing techniques for each problem. 
By exploiting the multi-view geometry explicitly, our proposed method can handle more general blur kernels that may result from camera rotations and forward motions.


%
%
%

\section{Related Works}
\label{sec:relatedWorks}

Few researchers have attempted to perform image matching on blurry images. 
Portz \etal~\cite{Portz2012} proposed an optical flow method that uses a blur-aware matching procedure originally introduced in tracking methods~\cite{Jin2005,Mei2008}.
Based on the assumed commutativity of blur operations, this method blurs the input images with the kernels of one another instead of deblurring them using their own kernels.

Lee \etal~\cite{Lee2011,Lee2013} extended this idea and proposed several methods for handling blurred input images in camera pose estimation~\cite{Lee2011} and dense stereo matching~\cite{Lee2013}. 
However, given that scene depth and camera motion can generate the exact blur kernels only when both values are correct, estimating these parameters separately would be inappropriate.
Moreover, the aforementioned works~\cite{Lee2013, Portz2012} are limited by a simple assumption that the blur kernel can be modeled by using linear optical flow vectors between consecutive frames.

By contrast, our proposed blur model (Section~\ref{sec:IPmodel}) covers more general camera motions by adopting the linear model in an Lie algebra \sethreeAlg space~\cite{Blanco2010}. 
The blur kernel is explicitly approximated by interpolating the camera path and depth maps between adjacent frames.
Figure~\ref{fig:blurModels} shows the difference between the conventional and proposed blur models.

Recent works~\cite{sellent2016stereo,zhen2016motion} have attempted to solve stereo matching and image deblurring jointly using the same blur model as the proposed one.
However, both of these methods depend on additional or external data.
The method proposed by Sellent \etal~\cite{sellent2016stereo} can only handle stereo video sequences, in  which per-frame depth cues are available.
Zhen and Stevenson~\cite{zhen2016motion} proposed a method for single-view image sequences, but this method requires additional data, including inertial measurements and sharp noisy frames.

Some methods solve the super-resolution and MVS problems in a single framework~\cite{Bhavsar2010, Lee2013_2}, which is shown to increase the accuracy of both the restored images and depth maps. 
However, the multi-frame super-resolution framework used in~\cite{Bhavsar2010, Lee2013_2} only works when accurate matching information is available in sub-pixel units.
Therefore, these methods cannot jointly estimate super-resolved images and depth maps for blurry input images because of the large errors in correspondences.

Some researchers proposed to solve super-resolution and deblurring jointly~\cite{takeda2011removing,bascle1996motion}.
The method proposed by Bascle \etal~\cite{bascle1996motion} relies on external tracking information to estimate the blur kernel using the trajectory of the object and establish the sub-pixel correspondence for multi-frame super-resolution.
However, this method is applicable only on some objects, which should be easy to track, not on the entire image.
Takeda and Milanfar~\cite{takeda2011removing} proposed an intriguing method to handle a spatio-temporal super-resolution and deblurring problem in a spatially invariant 3D deconvolution framework.
However, this method cannot handle large blur kernels because the size of motion vectors between consecutive frames is limited by a few pixels.



\section{Modeling Imaging Process}
\label{sec:IPmodel}
We examine an image sequence captured by a single moving camera where the target scene is assumed to be static to enable stereo matching and camera pose estimation.
In this study, an image is defined as a mapping that uses a 2D pixel coordinate vector as input and a 3D color vector as output (in the case of typical RGB images). 
The color value of the pixel ${\bf{x}}$ = ${\left[ {x,y} \right]^T}$ of image ${I}$ is given by $I\left( {\bf{x}} \right)$. When the query 2D coordinate has non-integer values, the color values are interpolated using the color values of the neighboring grid points. 
We apply bilinear interpolation throughout this paper.

In the following, the input images are denoted by ${B_t}$\rq{}s, with $t$ representing the time when the images are captured. 
An acquired image ${B_t}$  is assumed to be the accumulation of the sensor output from the opening (${t_o}$) to the closing (${t_c} = t$) of the camera shutters. 
We model this capturing process by assuming the presence of ideal clean and HR images during the shutter time. 
By denoting the ideal images at time $\tau$ as ${I_\tau}$, a real image ${B_t}$ is considered as the downsampled version of the integral of ${I_\tau}$ as follows:
\begin{equation}
{B_t} = \frac{1}{{{t_c} - {t_o}}}\left( {\int_{{t_o}}^{{t_c=t}} {{I_\tau}d\tau } } \right)\downarrow,
\label{eq:imageModel}
\end{equation}
where the down arrow represents the downsampling process.

%
A blur is generated for the static scene because of the camera movement during the shutter time.
Therefore, the images ${I_\tau}$\rq{}s change over time. 
However, the difference between ${I_\tau}$ and ${I_t}$, clean HR image at time $t$, is not large because such variation is caused by the slight camera motions that take place within a short period (less than ${t_c} - {t_o}$). 
Therefore, ${I_\tau}$ can be approximated by warping ${I_t}$, if the relative poses of the cameras and the scene structure are both known.

We denote the pose of the camera and the depth map of the image ${I_t}$ by ${\bf{P}}_{t}$ and ${D_t}$, respectively, and then denote the time-invariant camera intrinsic matrix by $\bf{K}$. Using these notations, the warping process is expressed as follows:
\begin{equation}
I_\tau \left( {\bf{x}} \right) \approx I_t\left( {{W^{\tau \to t}}\left( {\bf{x}} \right)} \right),
\label{eq:approxHtau}
\end{equation}
where the function $W^{\tau \to t}\left(  \cdot  \right)$ computes the warped pixel position from the camera ${\bf{P}}_{\tau}$ to ${\bf{P}}_{t}$, and can be expressed as follows:
\begin{align}
{{\bf{x}}_t} &= {W^{\tau \to t}}\left( {\bf{x}} \right) \\ \nonumber
 &= {i_2}\left( {{\bf{K}}{i_3}\left( {{{\left( {{{\bf{P}}_{t}}} \right)}^{ - 1}}{{\bf{P}}_{\tau}}{h_3}\left( {\frac{1}{{{D_\tau }\left( {\bf{x}} \right)}}{{\bf{K}}^{ - 1}}{h_2}\left( {\bf{x}} \right)} \right)} \right)} \right),
\label{eq:warpingFunction}
\end{align}
where functions ${i_2}\left(  \cdot  \right)$ and ${i_3}\left(  \cdot  \right)$ convert the homogeneous coordinates into inhomogeneous coordinates in the 2D and 3D spaces, respectively, while ${h_2}\left(  \cdot  \right)$ and ${h_3}\left(  \cdot  \right)$ convert the inhomogeneous coordinates into homogeneous coordinates in the same spaces.

The integral in Equation~(\ref{eq:imageModel}) is approximated using a finite sum of images. 
The insertion of the image warping Equation~(\ref{eq:approxHtau}) generates the following:
\begin{equation}
B_t \approx {\capOper_t} \circ I_t,
\label{eq:capturingProcess}
\end{equation}
\begin{equation}
\left( {{\capOper_t} \circ I} \right)\left( {\bf{x}} \right) = \left( {\frac{1}{M}\sum\limits_{m = 1}^M {I\left( {{W^{{\tau _m} \to t}}\left( {\bf{x}} \right)} \right)} } \right) \downarrow,
\label{eq:blurOperator}
\end{equation}
where ${\tau_m} = \left( {{m \mathord{\left/
 {\vphantom {m M}} \right.
 \kern-\nulldelimiterspace} M}} \right)\left( {{t_c} - {t_o}} \right) + {t_o}$ and $M$ controls the degree of discretization. 
We define ${\capOper_t}\left(  \cdot  \right)$ as the operator on a general image $I$ to approximate the degradation resulting from the capturing process at time $t$.
Figure~\ref{fig:blurModels} illustrates the concept of this blur operation.

In practice, the values of ${D_{{\tau_m}}}\rq{}s$ and ${{\bf{P}}_{{\tau_m}}}\rq{}s$ are approximated using $D_t$, ${\bf{P}}_t$, and ${\bf{P}}_{s}$, where $s$ represents the time of the previous frame. ${\bf{P}}_{{\tau_m}}$ is sampled from the interpolated camera path between ${\bf{P}}_t$ and ${\bf{P}}_{s}$. The interpolation is conducted in the Lie algebra \sethreeAlg space~\cite{Blanco2010}. Given $\Delta {{\bf{P}}_{t,s}} = \log \left( {{{\bf{P}}_t} \cdot {{\left( {{{\bf{P}}_{s}}} \right)}^{ - 1}}} \right)$, the interpolation is performed as follows:
\begin{equation}
{{\bf{P}}_{{\tau _m}}} = \exp \left( {\frac{{{\tau _m} - s}}{{t - s}} \cdot \Delta {{\bf{P}}_{t,s}}} \right) \cdot {{\bf{P}}_s},
\label{eq:interpolatingP}
\end{equation}
where $\log$ and $\exp$ denote the logarithmic and exponential maps between the Lie group \sethree space, where the actual camera pose matrices resides, and the Lie algebra \sethreeAlg space~\cite{Blanco2010}. 
Note that the proposed method might work unreliably when the camera motion between the consecutive frames is too complex to be approximated by the simple interpolation scheme in Equation~(\ref{eq:interpolatingP}), for example, when the camera vibrates with a frequency much higher than the camera frame rate.

After obtaining the camera pose at time $\tau_{m}$, the depth map ${D_{{\tau_m}}}$ can be approximated by warping the closest depth map $D_{t}$. The warped depth value can be computed by reprojecting $D_{t}$ to the world coordinate and then projecting this map the virtual camera at ${\bf{P}}_{\tau_{m}}$. The projected value is actually the depth of the point from ${\bf{P}}_{\tau_{m}}$. The capturing operator ${\capOper_t}\left(  \cdot  \right)$ is only dependent on $D_t$, ${\bf{P}}_t$, and ${\bf{P}}_{s}$.

\section{Unified Energy Formulation}
\label{sec:energyFormulation}
This study aims to estimate the latent images ${I_t}$\rq{}s with the corresponding depth maps ${D_t}$\rq{}s and camera poses ${{\mathbf{P}}_t}$\rq{}s from a blurred, LR image sequence, ${B_t}$\rq{}s. 
We assume that the intrinsic parameters ${\mathbf{K}}$ are previously known. 
Given that the target variables are interrelated, the proposed method estimates them altogether by optimizing a single unified energy function.

The total energy function $E$ is defined by the sum of energy functions, $E_t$, which is defined for each single frame at time $t$. 
$E_t$ comprises three terms, with each term having a unique physical meaning:
\begin{equation}
E = \sum\limits_t {{E_t}},
\label{eq:energyFunction}
\end{equation}
\begin{equation}
{E_t} = E_t^m + E_t^s + E_t^r,
\label{eq:energyFunctiont}
\end{equation}
where the matching, self-consistency, and regularization terms are presented from left to right.

\subsection{Matching term}
\label{sec:matchingTerm}
The first term relates the images from the consecutive frames based on the scene structure and camera motion. 
Given the static target scene, the images warped into a specific frame must coincide if the warping is based on correct depth maps and camera poses. 


In the proposed matching term, we match the input blurred LR image, $B_t$, with the latent images of the neighboring frames, $I_s$\rq{}s, where $s \in N\left( t \right)$ denotes the time index for the neighboring frames of $t$. 
Therefore, an additional one-way blur operation for matching is performed, where $I_s$\rq{}s must be blurred and downsampled by the capturing operator of $B_t$. The matching term is defined as follows:
\begin{equation}
E_t^m = \sum\limits_{s \in N\left( t \right)} {\sum\limits_{{\mathbf{x}} \in {\Omega _{ts}}} {{{\left\| {{B_t}\left( {\mathbf{x}} \right) - {\capOper _{t}} \circ {I_s}\left( {{W^{t \to s}}\left( {\mathbf{x}} \right)} \right)} \right\|}_1}} }.
\label{eq:matchingTerm}
\end{equation}
The matching term only considers the pixels in the set $\Omega_{ts}$, which represents the visible area of the image domain at time $t$ in terms of the camera at $s$. Section~\ref{sec:occHandling} discusses how this area is determined. 
We use L1-norm, which generates reliable results and is highly robust to the presence of noise and occlusion~\cite{Wedel2008}. 

In terms of MVS matching, the proposed methods try to determine the plausible scene structure and camera poses that satisfy the brightness constancy assumption by minimizing the matching term. The same matching term is also used as the evidence of super-resolution for restoring $I_s$\rq{}s from LR observations based on the estimation of the latent images.


\subsection{Self-consistency term}
The self-consistency term $E_t^s$ is derived from the imaging process in Equation~(\ref{eq:blurOperator}) as follows:
\begin{equation}
E_t^s = {\lambda _s}\sum\limits_{\mathbf{x}} {{{\left\| {{B_t}\left( {\mathbf{x}} \right) - {\capOper _{t}} \circ {I_t}\left( {\mathbf{x}} \right)} \right\|}_1}},
\label{eq:IPterm}
\end{equation}
which makes the solution consistent with the observation. 
Based on the depth maps and camera poses, the capturing operator $\capOper _{t}\left(  \cdot  \right)$ is constant and the equation is similar to the conventional data term in the extant deblurring methods. 
The parameter ${\lambda _s}$ controls the strength of this constraint.

\subsection{Regularization term}
Although the matching and self-consistency terms can compensate each other, they both rely on possibly noisy input images. 
The additional term regularizes the solutions to suppress the errors. 
In the proposed framework, we use typical total variation (TV) priors for the depth maps and latent images. 
Although originally introduced for denoising signals, TV priors has been frequently used in addressing image deblurring~\cite{Xu2010}, super-resolution~\cite{Farsiu2004}, and stereo matching problems~\cite{Ranftl2012}. 

The TV priors used in the proposed method is defined as follows:
\begin{equation}
E_t^r = {\lambda _d}\sum\limits_{\mathbf{x}} g_{t}\left( \bf{x} \right) {{{\left\| {\nabla {D_t}\left( {\mathbf{x}} \right)} \right\|}_2}}  + {\lambda _i}\sum\limits_{\mathbf{x}} {{{\left\| {\nabla I_t\left( {\mathbf{x}} \right)} \right\|}_2}},
\label{eq:regularizingTerm}
\end{equation}

where $\nabla I\left( {\mathbf{x}} \right)$ represents the gradient value of image $I$ at pixel $\mathbf{x}$. The weighting function  $g_{t}\left( \bf{x} \right)$ is used for edge-preserving smoothness with the same definition as proposed in~\cite{hyun2015generalized}. We use the magnitude of L2-norm to make the TV priors isotropic while preserving the discontinuities in the images and depth maps.
The parameters ${\lambda _d}$ and  ${\lambda _i}$ determine the degree of regularization on the depth maps and latent images, respectively.

\section{Optimization}
\label{sec:optimization}
The optimization of Equation~(\ref{eq:energyFunction}) is a complex process that serves as a function of many variables ($D_t$\rq{}s, ${\mathbf{P}}_t$\rq{}s, and $I_t$\rq{}s for all frames).
This process is also highly nonlinear because of the warping operations. Therefore, instead of obtaining the global optimum, we attempt to secure a favorable approximated solution by adopting several strategies. At the core of this solution is a divide-and-conquer strategy or an iterative and alternating optimization of variables. The proposed framework uses two-phase iterations in which the structures (cameras and depth maps) and latent images are alternatingly updated.

Algorithm~\ref{alg:optimization} presents the overall optimization procedure.
$T$ denotes the number of frames in the input image sequence, while $\textit{max\_iter}$ denotes the number of iterations set by users.
The solutions almost converge after three iterations, which is the chosen $\textit{max\_iter}$ value of the proposed method. 


\begin{algorithm}[t]
\caption{The overall optimization procedure}
\label{alg:optimization}
\begin{algorithmic}

\State \% initialization
\For{$t = 1$ {\bf{to}} $T$}
	\State Initialize $D_t$, ${\mathbf{P}}_t$ by minimizing Equation~(\ref{eq:initializationScene})
\EndFor
\State \% main loop
\For{$iteration = 1$ {\bf{to}} $\textit{max\_iter}$}
	\State \% first phase : update images
	\For{$t = 1$ {\bf{to}} $T$}
		\State update $I_t$ by minimizing Equation~(\ref{eq:imgUpdate})
	\EndFor
	\State \% second phase : update depths and cameras
	\State approximate Equation~(\ref{eq:energyFunction}) using Equation~(\ref{eq:taylorExpansion})
	\State update $D_t$'s and ${\mathbf{P}}_t$'s by using IRLS
\EndFor
\end{algorithmic}
\end{algorithm}

\begin{table*}[t]
\caption{The performance comparison of deblurring performance for synthetic datasets. All the PSNR(dB) values are averaged for the whole frames in each sequence.}
\centering
\resizebox{0.78\linewidth}{!}{
\begin{tabular}{l r r r r r}
	\toprule
	\textbf{Methods} & \textbf{Dolls} & \textbf{Reindeer} & \textbf{InteriorScene}~\cite{InteriorSceneURL} & \textbf{WorkDesk}~\cite{WorkDeskURL} & \textbf{avg.} \\
	\toprule
	Bicubic interpolation (Bic.) & 23.52 & 29.54 & 26.82 & 20.45 & 25.08\\
	Bic. + Lee and Lee\cite{Lee2013} & 11.17 & 22.52 & 15.19 & 10.88 & 14.94\\
	Timofte \etal~\cite{timofte2014} + Lee and Lee\cite{Lee2013} & 10.60 & 16.71 & 13.29 & 9.74 & 12.59\\
	Wang \etal~\cite{wang2015deep} + Lee and Lee\cite{Lee2013} & 11.07 & 21.44 & 15.08 & 10.86 & 14.61\\
	Bic. + Xu \etal~\cite{Jia2013} & 22.47 & 26.88 & 26.43 & 19.77 & 23.89\\
	Timofte \etal~\cite{timofte2014} + Xu \etal~\cite{Jia2013} & 19.68 & 22.66 & 23.52 & 17.71 & 20.89\\
	Wang \etal~\cite{wang2015deep} + Xu \etal~\cite{Jia2013} & 22.62 & 27.00 & 26.40 & 19.71 & 23.93\\
	Bic. + Kim and Lee\cite{hyun2015generalized} & 25.96 & 31.03 & 28.55 & 24.23 & 27.44\\
	Timofte \etal~\cite{timofte2014} + Kim and Lee\cite{hyun2015generalized} & 22.41 & 24.20 & 25.82 & 20.51 & 23.23\\
	Wang \etal~\cite{wang2015deep} + Kim and Lee\cite{hyun2015generalized} & 26.11 & 31.56 & 28.65 & 24.18 & 27.63\\
	Kim and Lee\cite{hyun2015generalized} + Wang \etal~\cite{wang2015deep} & 25.56 & 29.86 & 28.39 & 23.84 & 26.91\\
	Xu \etal~\cite{Jia2013} + Wang \etal~\cite{wang2015deep} & 21.24 & 24.10 & 24.82 & 18.33 & 22.12\\
	Proposed(w/o SR) + Bic. & 27.33 & 31.11 & 22.48 & 22.17 & 25.77\\
	Bic. + Proposed(w/o SR) & 26.92 & 30.97 & 27.73 & 24.71 & 27.58\\
	Proposed & \textbf{28.39} & \textbf{32.48} & \textbf{29.06} & \textbf{25.29} & \textbf{28.81}\\
	\bottomrule
\end{tabular}}
\label{tab:quantBlur}
\end{table*}

\begin{table}[t]
\caption{Depth and camera pose estimation performance of synthetic datasets. The errors are measured using PSNR and relative errors (rel.) for depth, and absolute trajectory error ($e_{ate}$) for pose~\cite{engel2016dso}. All errors are averaged for the whole frames in each sequence.}
\centering
\resizebox{0.99\linewidth}{!}{
\begin{tabular}{cl rr r}
	\toprule
	\multicolumn{1}{c}{\multirow{2}{*}{\textbf{Datasets}}} & \multicolumn{1}{c}{\multirow{2}{*}{\textbf{Methods}}} & \multicolumn{2}{c}{\textbf{Depth errors}} & \multicolumn{1}{c}{\textbf{Pose errors}} \\ 
	\cmidrule(lr){3-4} \cmidrule(lr){5-5} 
	\multicolumn{1}{c}{} & \multicolumn{1}{c}{} & \textbf{PSNR(dB)} & \textbf{rel.} & \textbf{traj.($e_{ate}$)} \\ 
	\toprule
	\multirow{6}{*}{\textbf{Dolls}~\cite{middlebury2005}} & Bic. + Lee \etal~\cite{Lee2011} & - & - & 0.1220\\ 
	 & Bic. + Lee and Lee~\cite{Lee2013} & 19.76 & 0.6700 & -\\ 
	 & Bic. + Baseline & 41.79 & 0.0560 & 0.0046\\ 
	 & \cite{wang2015deep} +~\cite{Jia2013} + Baseline & 40.51 & 0.0676 & 0.0028\\ 
	 & \cite{wang2015deep} +~\cite{hyun2015generalized} + Baseline & 41.70 & 0.0568 & 0.0078\\ 
	 & Proposed(w/o SR) + Bic. & 43.47 & 0.0396 & \textbf{0.0005}\\ 
	 & Bic. + Proposed(w/o SR) & 43.50 & 0.0375 & 0.0011\\ 
	 & Proposed & \textbf{45.37} & \textbf{0.0336} & 0.0027\\ 
	\midrule
	\multirow{6}{*}{\textbf{Reindeer}~\cite{middlebury2005}} & Bic. + Lee \etal~\cite{Lee2011} & - & - & 0.0107\\ 
	 & Bic. + Lee and Lee~\cite{Lee2013} & 23.00 & 0.4982 & -\\ 
	 & Bic. + Baseline & 37.79 & 0.1084 & 0.0021\\ 
	 & \cite{wang2015deep} +~\cite{Jia2013} + Baseline & 37.23 & 0.2026 & 0.0022\\ 
	 & \cite{wang2015deep} +~\cite{hyun2015generalized} + Baseline & 37.72 & 0.1099 & 0.0036\\ 
	 & Proposed(w/o SR) + Bic & 36.52 & 0.1321 & 0.0005\\ 
	 & Bic. + Proposed(w/o SR) & 37.41 & 0.1143 & \textbf{0.0005}\\ 
	 & Proposed & \textbf{37.99} & \textbf{0.1055} & 0.0012\\ 
	\midrule
	\multirow{6}{*}{\textbf{InteriorScene}~\cite{InteriorSceneURL}} & Bic. + Lee \etal~\cite{Lee2011} & - & - & 1.9355\\ 
	 & Bic. + Lee and Lee~\cite{Lee2013} & 23.15 & 0.4641 & -\\ 
	 & Bic. + Baseline & 30.82 & 0.1647 & 0.1548\\ 
	 & \cite{wang2015deep} +~\cite{Jia2013} + Baseline & 30.93 & 0.1627 & 0.1288\\ 
	 & \cite{wang2015deep} +~\cite{hyun2015generalized} + Baseline & 30.41 & 0.1812 & 0.0923\\ 
	 & Proposed(w/o SR) + Bic & 21.26 & 0.5253 & 0.0974\\ 
	 & Bic. + Proposed(w/o SR) & 30.19 & 0.1802 & \textbf{0.0281}\\ 
	 & Proposed & \textbf{31.28} & \textbf{0.1617} & 0.1461\\ 
	\midrule
	\multirow{6}{*}{\textbf{WorkDesk}~\cite{WorkDeskURL}} & Bic. + Lee \etal~\cite{Lee2011} & - & - & 2.8334\\ 
	 & Bic. + Lee and Lee~\cite{Lee2013} & 26.01 & 0.4411 & -\\ 
	 & Bic. + Baseline & 36.85 & 0.0949 & 0.1392\\ 
	 & \cite{wang2015deep} +~\cite{Jia2013} + Baseline & 36.23 & 0.1057 & 0.1950\\ 
	 & \cite{wang2015deep} +~\cite{hyun2015generalized} + Baseline & 30.82 & 0.2479 & 0.4953\\ 
	 & Proposed(w/o SR) + Bic & 36.16 & 0.1031 & 0.3481\\ 
	 & Bic. + Proposed(w/o SR) & \textbf{39.90} & \textbf{0.0544} & \textbf{0.0914}\\ 
	 & Proposed & 38.13 & 0.0781 & 0.5048\\ 
	\bottomrule
\end{tabular}}
\label{tab:quant}
\end{table}

\subsection{Update of the depth maps and camera poses}
\label{sec:optDepthAndCamera}
In the first phase of each iteration, we optimize the variables on the scene structure, $D_t$\rq{}s and ${\mathbf{P}}_t$\rq{}s, with the fixed latent images, $I_t$\rq{}s. 
The energy function then becomes similar to that of the variational framework for optical flow~\cite{Sun2010} and we follow the optimization strategy employed in ~\cite{Sun2010}. 
At each iteration of this iterative optimization, the functions in the L1-norm for Equations~(\ref{eq:matchingTerm}) and~(\ref{eq:IPterm}) are approximated using the first-order Taylor expansion at the current solution. 

The linear approximation is conducted by calculating the partial derivatives of the warping equation in terms of individual depth value and camera pose as parameterized by the six-dimensional vector on \sethreeAlg. 
Suppose that the current solution of our iterative algorithm lies at a point in the solution space, $D^0_t$, ${\mathbf{P}}^0_t$, and ${\mathbf{P}}^0_s$.
The backward image warping procedure from the frame at time $s$ to $t$ can be approximated as follows:
\begin{equation}
{I^0_s}\left( {\mathbf{x}} \right) = {\left. {I_s\left( {{W^{t \to s}}\left( {\mathbf{x}} \right)} \right)} \right|_{{D_t} = D_t^0,{P_t} = P_t^0,{P_s} = P_s^0}}, \nonumber
\end{equation}
\begin{equation}
\begin{array}{l}
I_s\left( {{W^{t \to s}}\left( {\mathbf{x}} \right)} \right)\\
 = {I_s^0}\left( {\mathbf{x}} \right) + \frac{{\partial I_s^0}}{{\partial {\mathbf{u}}}}\left( {\frac{{\partial {\mathbf{u}}}}{{\partial {D_t}\left( {\mathbf{x}} \right)}}\Delta {D_t}\left( {\mathbf{x}} \right) + \frac{{\partial {\mathbf{u}}}}{{\partial {\varepsilon _t}}}{\varepsilon _t} + \frac{{\partial {\mathbf{u}}}}{{\partial {\varepsilon _s}}}{\varepsilon _s}} \right),
\end{array}
\label{eq:taylorExpansion}
\end{equation}
where $\mathbf{u}$ is the warping-generated flow that serves as a function of the depth and camera parameters. 
The partial derivatives are actually Jacobians ~\cite{Blanco2010}.  

${\Delta {D_t}\left( {\mathbf{x}} \right)}$, $\epsilon_t$, and $\epsilon_s$ are variables that contribute to the solution as follows:
\begin{align}
{D_t}\left( {\mathbf{x}} \right) &= D_t^0\left( {\mathbf{x}} \right) + \Delta {D_t}\left( {\mathbf{x}} \right), \nonumber \\
{{\mathbf{P}}_t} &= \exp \left( {{\varepsilon _t}} \right){\mathbf{P}}_t^0, \nonumber \\
{{\mathbf{P}}_s} &= \exp \left( {{\varepsilon _s}} \right){\mathbf{P}}_s^0.
\label{eq:updateVariables}
\end{align}
Given that all terms in the L1-norm have been linearized, these variables can be efficiently estimated using the simple iteratively reweighted least square (IRLS) method~\cite{scales1988}. 

\subsection{Update of the latent images}
\label{sec:optLatentImages}
The latent images are optimized in the second phase of the outer loop.
The L1-norm functions for the target image $I_t$ in the matching term, Equation~(\ref{eq:matchingTerm}), provides information about the different blur and sampling of latent image $I_t$.
The self-consistency term in Equation~(\ref{eq:IPterm}) and the smoothness imposed by the regularization term in Equation~(\ref{eq:regularizingTerm}) are considered to provide a frame-by-frame representation of the energy function on $I_t$ as follows:
\begin{align}
\label{eq:imgUpdate}
\begin{array}{*{20}{l}}
  {\sum\limits_{s \in N\left( t \right)} {\sum\limits_{{\mathbf{x}} \in {\Omega _{ts}}} {{{\left\| {{\capOper_s} \circ {I_t}\left( {{W^{s \to t}}\left( {\mathbf{x}} \right)} \right) - {B_s}\left( {\mathbf{x}} \right)} \right\|}_1}} } } \\ 
  { + {\lambda _s}\sum\limits_{\mathbf{x}} {{{\left\| {{\capOper_t} \circ {I_t}\left( {\mathbf{x}} \right) - {B_t}\left( {\mathbf{x}} \right)} \right\|}_1}} } \\ 
  { + {\lambda _i}\sum\limits_{\mathbf{x}} {{{\left\| {\nabla {I_t}\left( {\mathbf{x}} \right)} \right\|}_2}} ,} 
\end{array}
\end{align}
which is optimized by finding the most plausible values that satisfy these competing constraints simultaneously.

We apply bilinear interpolation to sample the color values of non-grid points in image warping, and then apply simple box filtering for downsampling in the capturing operation.
This process makes the warping and capturing operations act as linear operators on the latent image after fixing the depth maps and camera poses. 
Consequently, the Equation~(\ref{eq:imgUpdate})denotes the sum of L1-norm and L2-norm on the linear functions of $I_t$ that can be easily optimized using IRLS~\cite{scales1988}.

\subsection{Initialization}
\label{sec:optInitialization}
We initialize the camera poses of the first two frames using a structure from motion software~\cite{wu2011visualsfm,wu2011multicore}.
After determining the camera poses of the first two frames, the depth maps $D_t$\rq{}s and remaining camera poses $P_t$\rq{}s can be initialized by sequentially minimizing the following equation frame-by-frame in a coarse-to-fine manner~\cite{Sun2010}:
\begin{align}
\label{eq:initializationScene}
E_t^{init} &= \sum\limits_{\mathbf{x}} {{{\left\| {\left( {{\capOper_t} \circ B_s} \right)\left( {{W^{t \to s}}\left( {\mathbf{x}} \right)} \right) - {\capOper_s} \circ B_t\left( {\mathbf{x}} \right)} \right\|}_1}} {\rm{ }} \nonumber \\
&+ {\lambda_d}\sum\limits_{\mathbf{x}} {{{\left\| {\nabla {D_t}\left( {\mathbf{x}} \right)} \right\|}_2}},
\end{align}
where $s$ denotes the time of the previous frame. Given that the estimated depth maps have LR, we upsample these maps to match the resolution of the target latent images, and then begin the main loop of the optimization. We adopt a simple bicubic interpolation method for the upsampling.

\begin{figure*}[t]
	\centering
	\begin{subfigure}[b]{0.243\textwidth}
		\includegraphics[width=\textwidth]{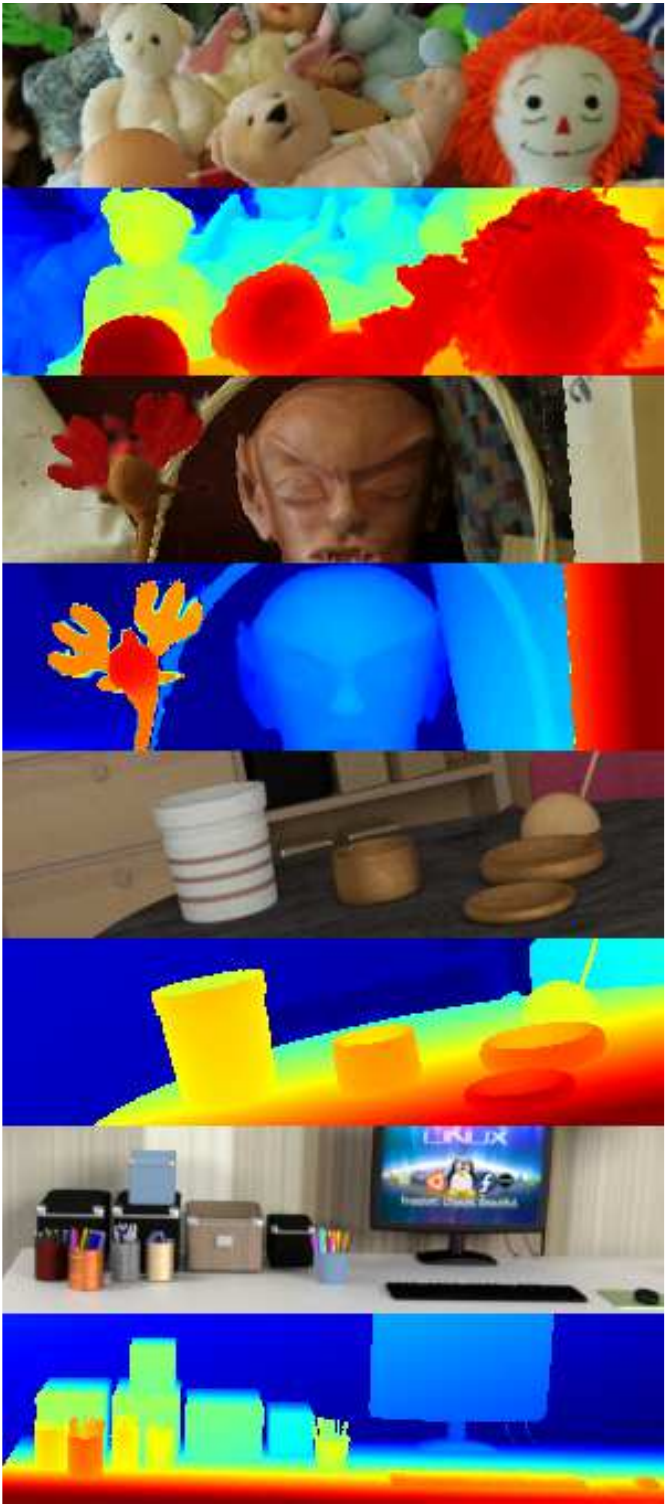}
		\caption{Ground truth}
	\end{subfigure}
	\begin{subfigure}[b]{0.243\textwidth}
		\includegraphics[width=\textwidth]{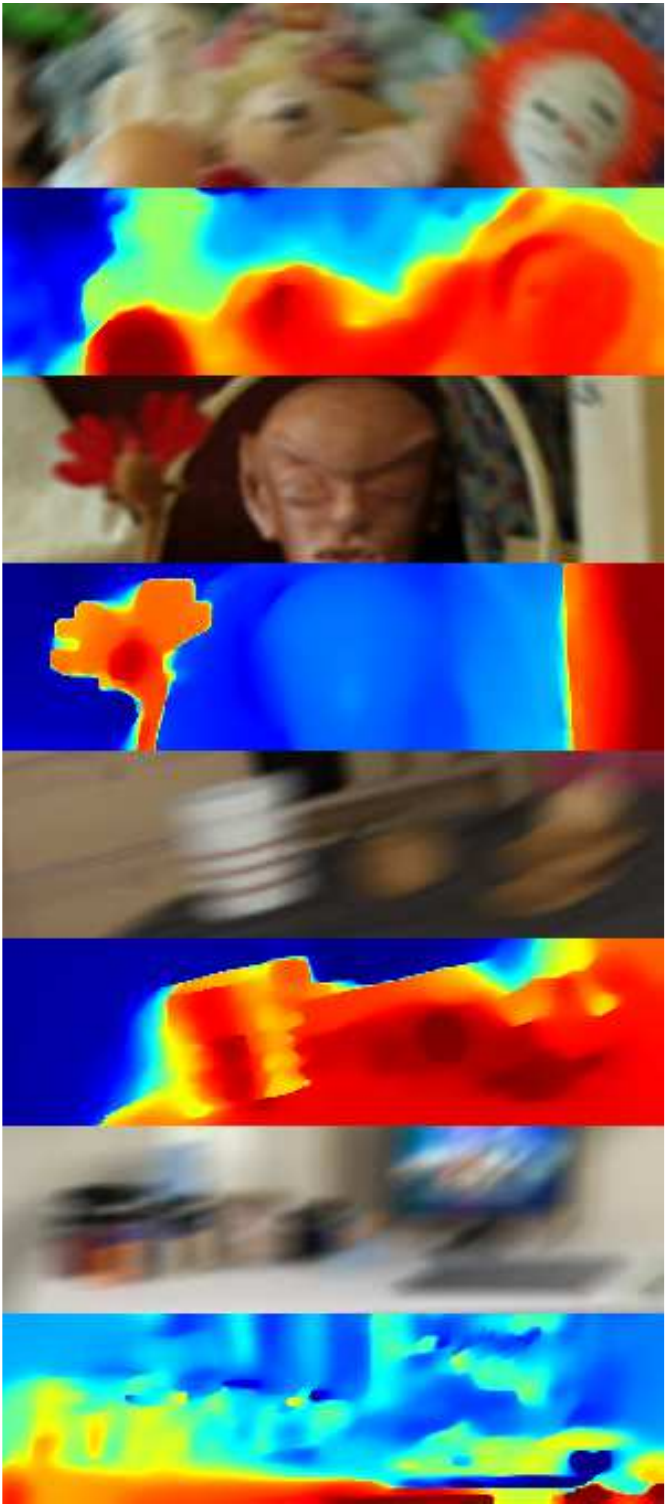}
		\caption{Bic. + baseline}
	\end{subfigure}
	\begin{subfigure}[b]{0.243\textwidth}
		\includegraphics[width=\textwidth]{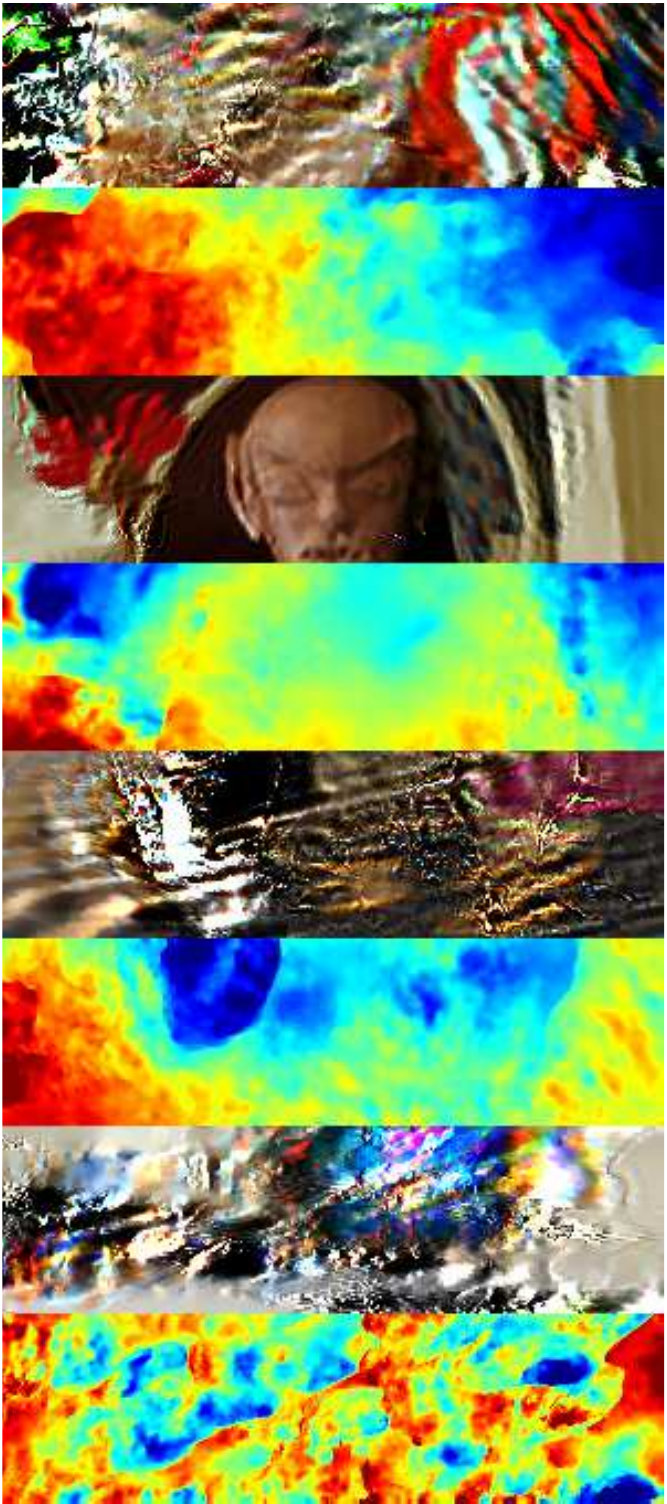}
		\caption{Bic.  +~\cite{Lee2013}}
	\end{subfigure}
	\begin{subfigure}[b]{0.243\textwidth}
		\includegraphics[width=\textwidth]{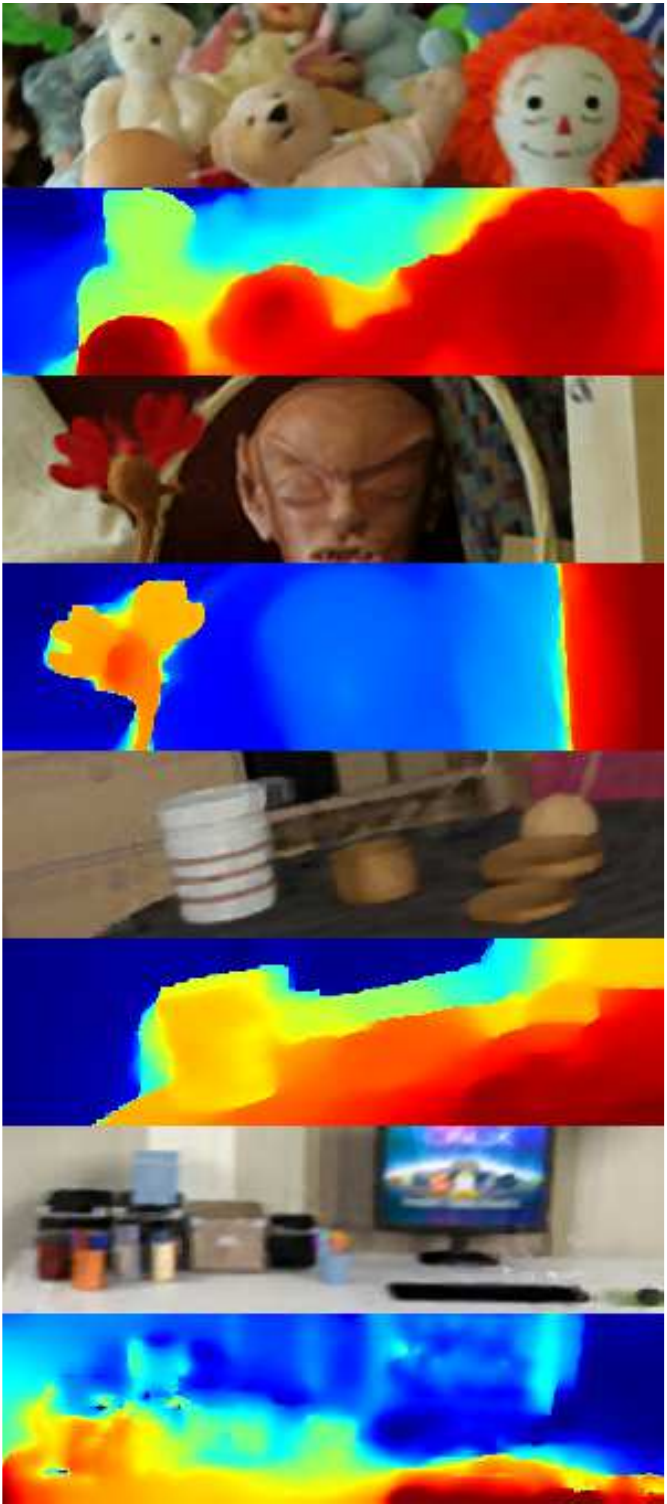}
		\caption{Proposed}
	\end{subfigure}
\caption{Comparison of the depth maps and latent images of synthetic datasets. 
Each pair of rows shows results on \textbf{Dolls}~\cite{middlebury2005},  \textbf{Reindeer}~\cite{middlebury2005}, \textbf{InteriorScene}~\cite{InteriorSceneURL}, and \textbf{WorkDesk}~\cite{WorkDeskURL} dataset from top to bottom . }
\label{fig:synthetic}
\end{figure*}

\subsection{Occlusion Handling}
\label{sec:occHandling}
Although the use of L1-norm for the matching term in Equation~(\ref{eq:matchingTerm}) makes the proposed method robust to existence of occlusion, modeling the visible area in ${\Omega _{ts}}$ can help generate precise depth values around the discontinuities.
Therefore, we update the visible area ${\Omega _{ts}}$ whenever the depth maps and camera poses are updated. Given the updated depth maps and camera poses, we update ${\Omega _{ts}}$ as follows:
\begin{equation}
\label{eq:occUpdate}
{\Omega _{ts}} = \left\{ {{\mathbf{x}}\left| {{D_t}\left( {\mathbf{x}} \right) > {D_t}\left( {\mathbf{y}} \right),\forall {\mathbf{y}} \in {\Theta _{ts}}\left( {\mathbf{x}} \right)} \right.} \right\},
\end{equation}
where ${\Theta _{ts}}$ represents the set of pixels in the camera at time $t$ that fall in the same area after warping.
\begin{equation}
\label{eq:occSet}
{\Theta _{ts}}\left( {\mathbf{x}} \right) = \left\{ {{\mathbf{y}}\left| {\left| {{W^{t \to s}}\left( {\mathbf{y}} \right) - {W^{t \to s}}\left( {\mathbf{x}} \right)} \right| \le 0.5} \right.} \right\}.
\end{equation}

\section{Experimental Results}
\label{sec:experimentalResults}
We test the validity of our proposed method on synthetic and real datasets. For comparison, we use the simple variational matching method as the baseline. This method solves the same optimization problem as the proposed method, except that the capturing operations are missed in the energy terms and the images are fixed to input images.


The values of some parameters are empirically determined. 
The proposed algorithm converges to favorable solutions when $\textit{max\_iter}$ is $3$ and $M$ is $50$. 
We use a large value of $\lambda_s$ ($30$) for all datasets to provide strong constraints on the solutions. 
We tune the value of $\lambda_d$ between $8$ to $10$ and the value of $\lambda_i$ between $0.3$ to $0.6$ based on the dataset. 
We set the upscale factor of the method to  $2$.

Our proposed framework is limited by its computational complexity. Specifically, we spend approximately five hours to process $10$ frames of $320\times240$ images in our Matlab implementation using a quad-core 3.2GHz CPU. 
This complexity may be addressed by running many parts of the algorithm on GPU in parallel.

\begin{figure*}[t]
	\centering
	\begin{subfigure}[b]{0.195\textwidth}
		\includegraphics[width=\textwidth]{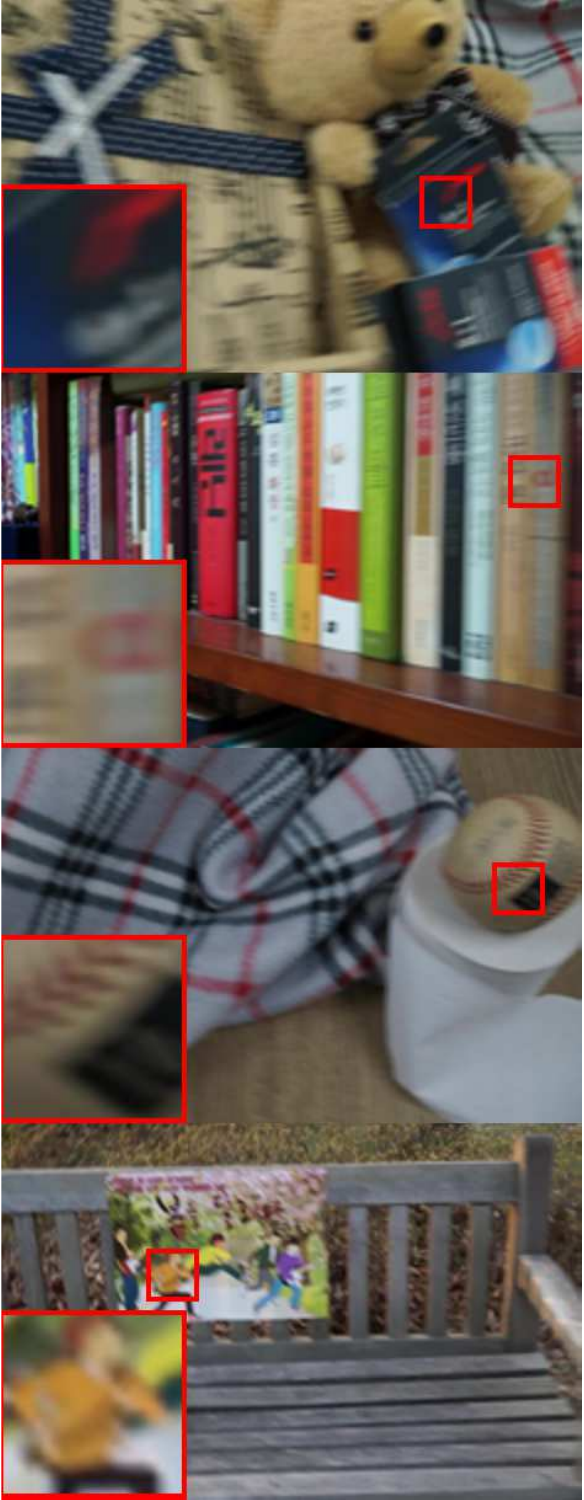}
		\caption{Bicubic interpolation}
	\end{subfigure}
	\begin{subfigure}[b]{0.195\textwidth}
		\includegraphics[width=\textwidth]{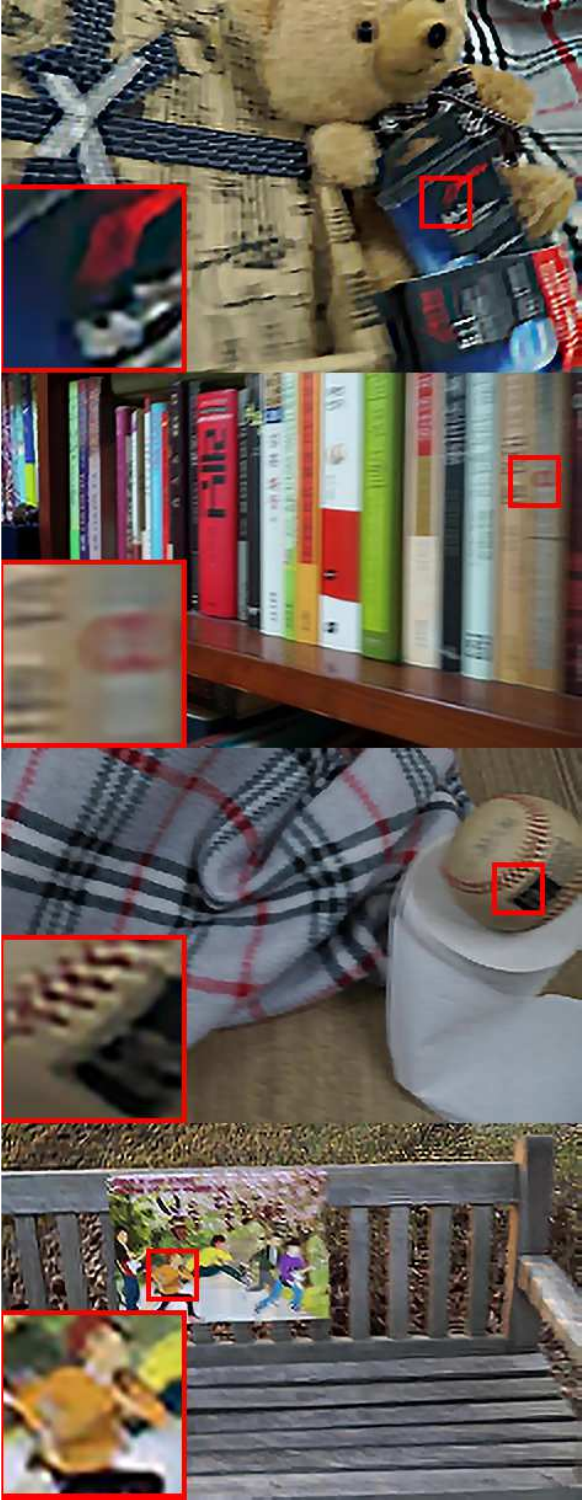}
		\caption{\cite{Jia2013} +~\cite{timofte2014}}
		\label{fig:realB}
	\end{subfigure}
	\begin{subfigure}[b]{0.195\textwidth}
		\includegraphics[width=\textwidth]{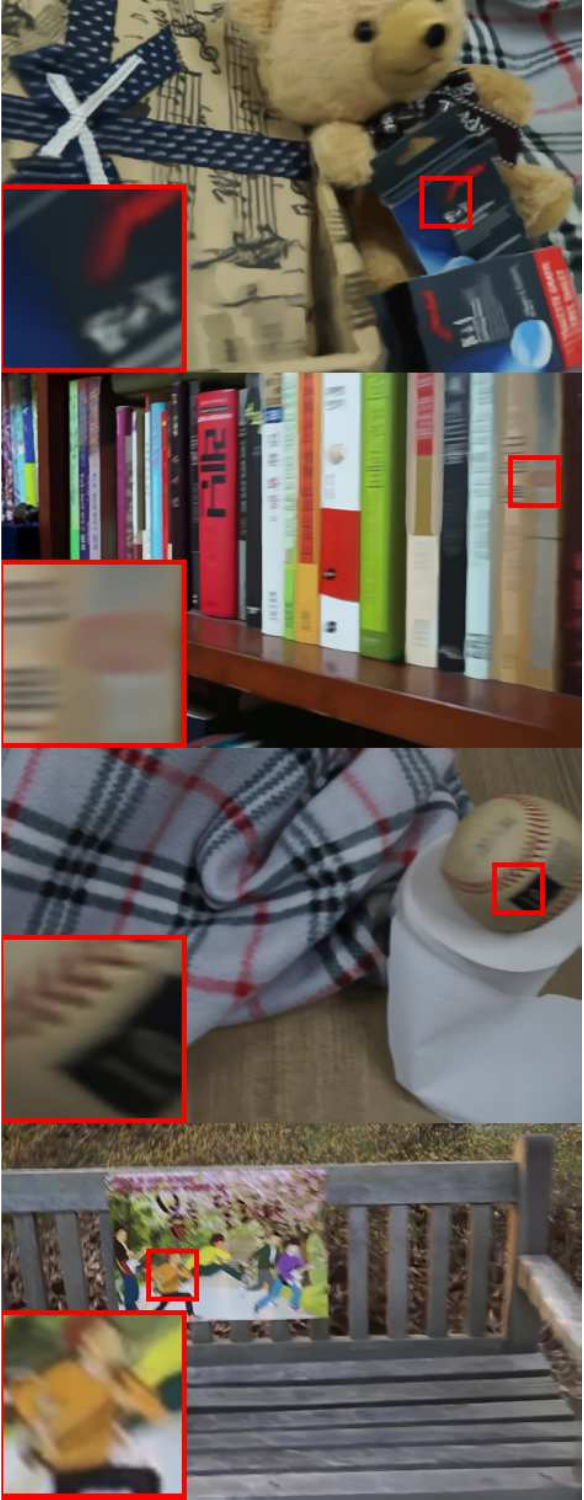}
		\caption{HR images +~\cite{cho2012}}
	\end{subfigure}
	\begin{subfigure}[b]{0.195\textwidth}
		\includegraphics[width=\textwidth]{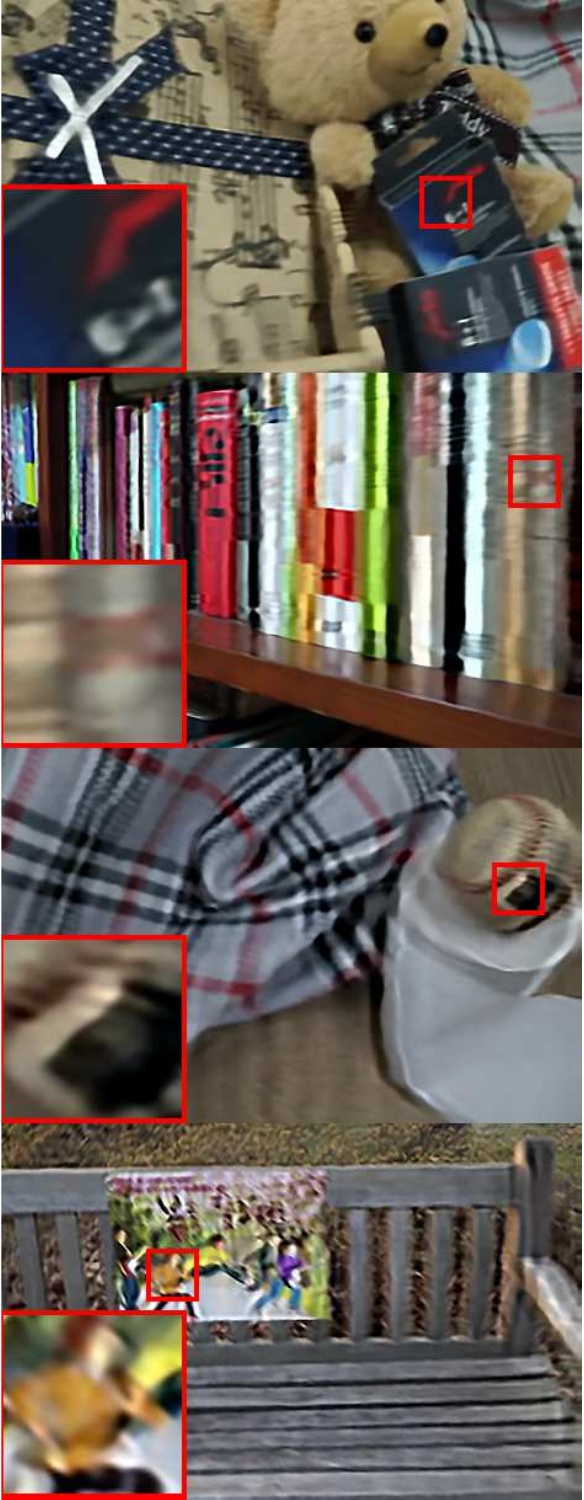}
		\caption{\cite{wang2015deep} +~\cite{hyun2015generalized}}
	\end{subfigure}
	\begin{subfigure}[b]{0.195\textwidth}
		\includegraphics[width=\textwidth]{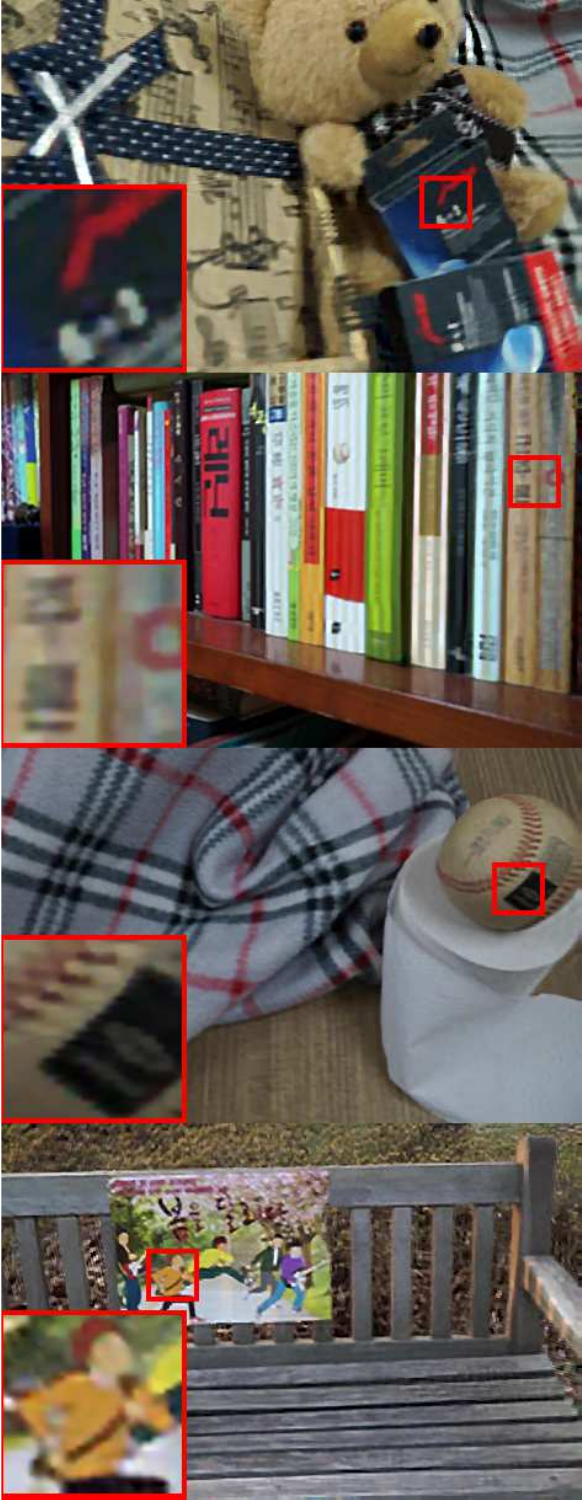}
		\caption{Proposed}
	\end{subfigure}
\caption{Comparison of the deblurring results on real datasets. }
\label{fig:realDeblur}
\end{figure*}

\subsection{Synthetic datasets}
No public datasets provide blurry LR images with corresponding ground-truth latent images, depth maps, and camera poses.
The desired datasets can be generated by synthesizing a simulated blur sequence using the Middlebury stereo datasets ~\cite{middlebury2005}. 
Given two images with ground-truth depth maps, the images between these two viewpoints are interpolated by assuming an imaginary camera path between the two reference views. Afterward, with a preset imaginary shutter time, the blurry images in each frame are approximated by summing up the intermediate images while the shutters are open. Similarly, we can generate a more realistic dataset using Blender~\cite{blender}. 
The intermediate images for these datasets are accurately rendered by using full 3D models. Figure~\ref{fig:synthetic} presents examples of synthesized datasets with their corresponding experimental results.

Table~\ref{tab:quant} and Table~\ref{tab:quantBlur} present the quantitative comparison results.
Table~\ref{tab:quant} shows the quantitative results of depth and camera pose estimation, while Table~\ref{tab:quantBlur} compares the deblurring results in terms of peak signal-to-noise ratio (PSNR).
The depth and image estimation errors are measured by comparing the reconstructed results with the closest intermediate sharp ground-truth ones (followed by scaling to address scale ambiguity for the case of depth maps). When we compute the depth errors for \textbf{Dolls} and \textbf{Reindeer} datasets~\cite{middlebury2005}, we cropped the depth maps to be the 70\% of original size at image center to ignore invalid regions around image boundary introduced by warping.

The third, fourth, and fifth rows of Table~\ref{tab:quant} show that using per-frame super-resolution and deblurring independently before matching may degrade the stereo matching performance as expected.
The method proposed by Lee and Lee~\cite{Lee2013} and Lee \etal.~\cite{Lee2011} performs worse than the baseline despite employing blur-aware matching. 
This result may be attributed to the fact that the degree of blur in our experiments is much more severe than that in the datasets used in~\cite{Lee2011,Lee2013}
and, furthermore, the scene structures in our datasets are more complex than the nearly planar structures in~\cite{Lee2011,Lee2013}.
The pose estimation performance of the proposed method seems less impressive compared to the depth estimation performance.
However, comparing the ground-truth trajectory to the restored camera trajectory itself can be problematic for blurry input images because it is ambiguous to specify a camera pose during the shutter time, especially when the the size of blur kernel is large as in \textbf{InteriorScene}~\cite{InteriorSceneURL} and \textbf{WorkDesk}~\cite{WorkDeskURL} datasets.
By contrast, the depth estimation errors are more significant because we can find the closest intermediate ground-truth depth map for a estimated depth map without ambiguity.


Table~\ref{tab:quantBlur} shows that the proposed method outperforms the combination of super-resolution methods~\cite{timofte2014,wang2015deep} and deblurring method~\cite{Lee2013,Jia2013,hyun2015generalized}, which implies that jointly estimating inter-related problems is effective in terms of image restoration.
The method proposed in~\cite{hyun2015generalized} also joinlty estimates the pixel correspondences (optical flow) and latent images (deblurring) from a video sequence.
Still, Table~\ref{tab:quantBlur} shows that leveraging multi-view constraints for deblurring problem and jointly solving it with super-resolution is more profitable.

The use of super-resolution clearly improves the accuracy of image restoration and depth estimation except for the case of \textbf{WorkDesk}~\cite{WorkDeskURL} dataset.
The surfaces in this dataset are weakly-textured and repeated, making the pixel-wise matching and super-resolution results less reliable.

\subsection{Real datasets}
For the real datasets, we use the proposed approach in~\cite{Zhang1999} for camera calibration. 
The shutter time and frames per second (FPS), both of which are necessary for interpolating the camera path and for simulating blurs for each frame, are obtained as metadata by taking an image sequence using commercial cameras.


Figure~\ref{fig:realDeblur} presents the comparison results of our proposed approach with those of other image restoration methods~\cite{Jia2013,timofte2014,cho2012,wang2015deep,hyun2015generalized}. 
Given that the images are blurred by real camera motions, we generate LR images by downsampling them manually to compare the super-resolution performances of these methods. 
The proposed method clearly outperforms the others even if the results of~\cite{cho2012} are obtained using the original HR images.
Some characters become recognizable and the textures representing the materials in the scene are well-restored in our results. 
Figure~\ref{fig:realB} also shows that performing the super-resolution after deblurring results in exaggeration of undesired artifacts, explaining the low PSNR values of 'super-resolution after deblurring' approaches in Table~\ref{tab:quantBlur}.
\section{Conclusion}
\label{sec:conclusion}
We proposed a pioneering framework for jointly solving four inter-related computer vision problems, including dense depth reconstruction, camera pose estimation, super-resolution, and deblurring. We jointly modeled these problems using an energy function that is derived by revisiting the blurry image formulation. Our model allows more general camera motions and nonlinear blur kernels than the previously proposed blur-aware matching methods. Our experiments show that the proposed method outperforms the other related methods that only address one or two target problems in terms of depth maps and latent images.
%

{\small
\bibliographystyle{ieee}
\bibliography{egbib}

\begin{thebibliography}{10}\itemsep=-1pt

\bibitem{InteriorSceneURL}
\url{http://www.blendswap.com/blends/view/72340/}.

\bibitem{WorkDeskURL}
\url{http://www.blendswap.com/blends/view/69052/}.

\bibitem{blender}
\url{http://www.blender.org/}.

\bibitem{bascle1996motion}
B.~Bascle, A.~Blake, and A.~Zisserman.
\newblock Motion deblurring and super-resolution from an image sequence.
\newblock {\em \comECCV}, pages 571--582, 1996.

\bibitem{Bhavsar2010}
A.~Bhavsar and A.~Rajagopalan.
\newblock Resolution enhancement in multi-image stereo.
\newblock {\em IEEE Transactions on Pattern Analysis and Machine Intelligence},
  32(9):1721--1728, Sept 2010.

\bibitem{Blanco2010}
J.-L. Blanco.
\newblock A tutorial on se(3) transformation parameterizations and on-manifold
  optimization.
\newblock Technical report, University of Malaga, Sept. 2010.

\bibitem{cho2012}
S.~Cho, J.~Wang, and S.~Lee.
\newblock Video deblurring for hand-held cameras using patch-based synthesis.
\newblock {\em ACM Transactions on Graphics (TOG)}, 31(4):64, 2012.

\bibitem{engel2016dso}
J.~Engel, V.~Koltun, and D.~Cremers.
\newblock Direct sparse odometry.
\newblock In {\em arXiv:1607.02565}, July 2016.

\bibitem{Farsiu2004}
S.~Farsiu, M.~Robinson, M.~Elad, and P.~Milanfar.
\newblock Fast and robust multiframe super resolution.
\newblock {\em IEEE Transactions on Image Processing}, 13(10):1327--1344, Oct
  2004.

\bibitem{middlebury2005}
H.~Hirschmuller and D.~Scharstein.
\newblock Evaluation of cost functions for stereo matching.
\newblock In {\em \comCVPR}, pages 1--8, June 2007.

\bibitem{hyun2015generalized}
T.~Hyun~Kim and K.~Mu~Lee.
\newblock Generalized video deblurring for dynamic scenes.
\newblock In {\em \comCVPR}, pages 5426--5434, 2015.

\bibitem{Jin2005}
H.~Jin, P.~Favaro, and R.~Cipolla.
\newblock Visual tracking in the presence of motion blur.
\newblock In {\em \comCVPR}, volume~2, pages 18--25 vol. 2, June 2005.

\bibitem{Lee2011}
H.~S. Lee, J.~Kwon, and K.~M. Lee.
\newblock Simultaneous localization, mapping and deblurring.
\newblock In {\em \comICCV}, pages 1203--1210, Nov 2011.

\bibitem{Lee2013}
H.~S. Lee and K.~M. Lee.
\newblock Dense 3d reconstruction from severely blurred images using a single
  moving camera.
\newblock In {\em \comCVPR}, pages 273--280, June 2013.

\bibitem{Lee2013_2}
H.~S. Lee and K.~M. Lee.
\newblock Simultaneous super-resolution of depth and images using a single
  camera.
\newblock In {\em \comCVPR}, pages 281--288, June 2013.

\bibitem{Mei2008}
C.~Mei and I.~Reid.
\newblock Modeling and generating complex motion blur for real-time tracking.
\newblock In {\em \comCVPR}, pages 1--8, June 2008.

\bibitem{Portz2012}
T.~Portz, L.~Zhang, and H.~Jiang.
\newblock Optical flow in the presence of spatially-varying motion blur.
\newblock In {\em \comCVPR}, pages 1752--1759, June 2012.

\bibitem{Ranftl2012}
R.~Ranftl, S.~Gehrig, T.~Pock, and H.~Bischof.
\newblock Pushing the limits of stereo using variational stereo estimation.
\newblock In {\em IEEE Intelligent Vehicles Symposium}, pages 401--407, June
  2012.

\bibitem{scales1988}
J.~A. Scales and A.~Gersztenkorn.
\newblock Robust methods in inverse theory.
\newblock {\em Inverse problems}, 4(4):1071, 1988.

\bibitem{middlebury2003}
D.~Scharstein and R.~Szeliski.
\newblock High-accuracy stereo depth maps using structured light.
\newblock In {\em \comCVPR}, volume~1, pages I--195--I--202 vol.1, June 2003.

\bibitem{sellent2016stereo}
A.~Sellent, C.~Rother, and S.~Roth.
\newblock Stereo video deblurring.
\newblock In {\em \comECCV}, pages 558--575. Springer, 2016.

\bibitem{Sun2010}
D.~Sun, S.~Roth, and M.~Black.
\newblock Secrets of optical flow estimation and their principles.
\newblock In {\em \comCVPR}, pages 2432--2439, June 2010.

\bibitem{takeda2011removing}
H.~Takeda and P.~Milanfar.
\newblock Removing motion blur with space--time processing.
\newblock {\em IEEE Transactions on Image Processing}, 20(10):2990--3000, 2011.

\bibitem{timofte2014}
R.~Timofte, V.~De~Smet, and L.~Van~Gool.
\newblock A+: Adjusted anchored neighborhood regression for fast
  super-resolution.
\newblock In {\em IEEE Asian Conference on Computer Vision}, 2014.

\bibitem{wang2015deep}
Z.~Wang, D.~Liu, J.~Yang, W.~Han, and T.~Huang.
\newblock Deep networks for image super-resolution with sparse prior.
\newblock In {\em \comICCV}, pages 370--378, 2015.

\bibitem{Wedel2008}
A.~Wedel, T.~Pock, C.~Zach, D.~Cremers, and H.~Bischof.
\newblock An improved algorithm for {TV-L1} optical flow.
\newblock In {\em Proc. of the Dagstuhl Motion Workshop}, LNCS. Springer,
  September 2008.

\bibitem{wu2011visualsfm}
C.~Wu.
\newblock Visualsfm: A visual structure from motion system.
\newblock \url{http://ccwu.me/vsfm/}, 2011.

\bibitem{wu2011multicore}
C.~Wu, S.~Agarwal, B.~Curless, and S.~M. Seitz.
\newblock Multicore bundle adjustment.
\newblock In {\em \comCVPR}, pages 3057--3064. IEEE, 2011.

\bibitem{Xu2010}
L.~Xu and J.~Jia.
\newblock Two-phase kernel estimation for robust motion deblurring.
\newblock In {\em \comECCV}, ECCV'10, pages 157--170, Berlin, Heidelberg, 2010.
  Springer-Verlag.

\bibitem{Jia2013}
L.~Xu, S.~Zheng, and J.~Jia.
\newblock Unnatural l0 sparse representation for natural image deblurring.
\newblock In {\em \comCVPR}, pages 1107--1114, June 2013.

\bibitem{Zhang1999}
Z.~Zhang.
\newblock Flexible camera calibration by viewing a plane from unknown
  orientations.
\newblock In {\em \comICCV}, volume~1, pages 666--673 vol.1, 1999.

\bibitem{zhen2016motion}
R.~Zhen and R.~L. Stevenson.
\newblock Motion deblurring and depth estimation from multiple images.
\newblock In {\em Image Processing (ICIP), 2016 IEEE International Conference
  on}, pages 2688--2692. IEEE, 2016.

\end{thebibliography}
}

\end{document}